\documentclass[conference]{IEEEtran}

\IEEEoverridecommandlockouts
\usepackage{cite}
\usepackage{amsmath,amssymb,amsfonts}
\usepackage{algorithmic}
\usepackage{graphicx}
\usepackage{balance}
\usepackage{textcomp}
\def\BibTeX{{\rm B\kern-.05em{\sc i\kern-.025em b}\kern-.08em
    T\kern-.1667em\lower.7ex\hbox{E}\kern-.125emX}}

\usepackage{hyperref} 
\usepackage{booktabs} 
\usepackage{multirow} 
\usepackage{longtable} 
\usepackage{rotating} 
\usepackage[table]{xcolor} 

\makeatletter
\newcommand{\linebreakand}{
  \end{@IEEEauthorhalign}
  \hfill\mbox{}\par
  \mbox{}\hfill\begin{@IEEEauthorhalign}
}
\makeatother

\begin{document}

\title{Adaptive Pruning of Deep Neural Networks for Resource-Aware Embedded Intrusion Detection on the Edge\thanks{This work has been submitted to an IEEE conference for possible publication. Copyright may be transferred without notice, after which this version may no longer be accessible.}}

\author{\IEEEauthorblockN{Alexandre Broggi}
\IEEEauthorblockA{\textit{Computer Science} \\
\textit{University of Massachusetts Dartmouth}\\
Dartmouth, Massachusetts, USA \\
abroggi@umassd.edu}
\and
\IEEEauthorblockN{Nathaniel Bastian}
\IEEEauthorblockA{
\textit{United States Military Academy}\\
USA \\
nathaniel.bastian@westpoint.edu}
\linebreakand
\IEEEauthorblockN{Lance Fiondella}
\IEEEauthorblockA{
\textit{University of Massachusetts}\\
Dartmouth, Massachusetts, USA \\
lfiondella@umassd.edu}
\and
\IEEEauthorblockN{Gokhan Kul}
\IEEEauthorblockA{\textit{Computer Science} \\
\textit{University of Massachusetts Dartmouth}\\
Dartmouth, Massachusetts, USA \\
gkul@umassd.edu}
}

\maketitle

\begin{abstract}
Artificial neural network pruning is a method in which artificial neural network sizes can be reduced while attempting to preserve the predicting capabilities of the network. This is done to make the model smaller or faster during inference time. In this work we analyze the ability of a selection of artificial neural network pruning methods to generalize to a new cybersecurity  dataset utilizing a simpler network type than was designed for. We analyze each method using a variety of pruning degrees to best understand how each algorithm responds to the new environment. This has allowed us to determine the most well fit pruning method of those we searched for the task. Unexpectedly, we have found that many of them do not generalize to the problem well, leaving only a few algorithms working to an acceptable degree.
\end{abstract}


\section{Introduction}
In recent years, Artificial Neural Networks (ANNs) have gained significant popularity due to their impressive capabilities in solving complex tasks across various domains, particularly in natural language processing~\cite{fanni2023natural,zhang2020task}, computer vision~\cite{zhang2020task}, and cybersecurity~\cite{ACIIOTdataset,mirsky2018kitsune}.
However, the increasing size and complexity of ANN models present practical challenges, particularly regarding their computational and memory demands~\cite{xu2020bert,he2023structured}.

Internet of Battlefield Things (IoBT)~\cite{joshi2023applications} applications heavily depend on portable devices that need to be embedded or carried along with moving operations centers at the edge~\cite{milcom}. Applications of Artificial Intelligence (AI) and ANN can prove to be extremely useful in IoBT settings. However, due to the heavy computational and memory demands, containing ANNs in chips that can be embedded is challenging. This motivates the exploration of efficient model optimization techniques, one of which is neural network pruning \cite{Distangling,he2023structured}.

ANNs~\cite{lippmann1988introduction} are computational models inspired by the structure and function of the human brain, comprising interconnected layers of artificial neurons that perform mathematical operations to learn and predict relationships in data. These models are commonly trained using optimization techniques like gradient descent, which iteratively minimizes a loss function to improve prediction accuracy. Through the use of activation functions and multiple layers, ANNs can approximate highly complex functions, enabling them to model nonlinear relationships in data. Despite their impressive power, modern ANN architectures, especially deep learning models, have grown exponentially in size, involving billions of parameters that require unreasonable computational resources for training and for inference~\cite{han2015learning}.

To address these scalability concerns, a technique known as neural network pruning has been developed~\cite{he2023structured}. Pruning involves reducing the size of a trained model by systematically removing neurons or weights that contribute the least to the model's performance. This results in a smaller, more efficient model with reduced storage requirements and faster inference times, while retaining much of the original predictive accuracy. Pruning methods can generally be classified into several categories: structured pruning~\cite{he2023structured}, unstructured pruning~\cite{han2015learning}, and hybrid approaches~\cite{LEE2022107988}. Each of these techniques aims to achieve an optimal balance between model efficiency and performance. 

The problem of model pruning is especially relevant in resource-constrained environments, such as Internet of Things (IoT) and IoBT applications, where computational efficiency and size of the model are crucial, so that these models can be used on the edge and even be embedded on hardware such as memristor CMOS designs. Therefore, in the context of network security, deploying lightweight, efficient models is essential for real-time intrusion detection, where large, high latency models may be impractical. The ACI IoT dataset~\cite{ACIIOTdataset} serves as a benchmark in this domain, representing real-world IoT traffic and providing a suitable challenge for evaluating the effectiveness of different pruning strategies. 

This paper aims to survey the landscape of neural network pruning mechanisms, with a focus on evaluating their efficacy on an intrusion detection benchmark dataset. Specifically, we compare multiple pruning methods applied to a deep neural network trained on the ACI IoT dataset, analyzing their effects on model performance and computational efficiency. We also propose  an alternate approach to an existing pruning mechanism, BERT-Theseus~\cite{xu2020bert}, 
to enhance its effectiveness for network intrusion detection. By doing so, we aim to provide a deeper understanding of how pruning can contribute to efficient deep learning models in the context of cybersecurity.

The remainder of this paper is organized as follows. Section~\ref{sec:Pruning} reviews the relevant literature on neural network pruning techniques, including structured, unstructured, and hybrid approaches. Section~\ref{sec:Algorithms} describes the algorithms we will be using. Section~\ref{sec:Setup} describes the experimental setup, including the architecture of the model used, the dataset, and the implementation details. Section~\ref{sec:Results} presents the results, comparing the impact of different pruning strategies on model accuracy and computational efficiency. Finally, Section~\ref{sec:Conclusion} concludes the paper with a discussion of our findings and potential future research directions.

Concretely, the contributions of this paper are as follows:
\begin{itemize}
\item A comprehensive survey of pruning approaches, highlighting their respective strengths and weaknesses when applied to the ACI IoT dataset.
\item An empirical comparison of different pruning techniques on a deep neural network for network intrusion detection.
\item Proposed Iterative Theseus, improving on an existing pruning mechanism, Bert-Theseus~\cite{xu2020bert}, demonstrating enhanced performance in terms of model efficiency and accuracy.
\end{itemize}

In the context of ever-growing model sizes and the need for efficient deployment, neural network pruning remains a critical area of research. The results presented in this study aim to guide researchers and practitioners towards more efficient deployment of deep learning models in security-sensitive applications.

\section{Pruning}
\label{sec:Pruning}

Pruning is a method of model compression for energy efficiency, training time efficiency or space requirements. Although pruning is not the only method that can accomplish this~\cite{Distangling}, as you could also use quantization or knowledge distillation~\cite{hinton2015distillingknowledgeneuralnetwork}, pruning is the most straightforward. As a method, Quantization~\cite{gong2014compressing} is the use of computationally smaller numbers to perform calculations, such as using \texttt{int32} instead of \texttt{int64} for numbers, using smaller quanta. This allows the model to be in a smaller space and to be mathematically faster to calculate but can be tricky to deal with given the loss in precision. Knowledge distillation is the method by which a large ANN model is used to generate more efficient training for a smaller ANN model. The core concept is that the larger ANN has taken time to understand correlations between different items and those specific correlations can be transferred to smaller models easier then it was to find them on the smaller model alone. On the other hand, pruning is the method of removing layers, filters, or weights from a model that are less important or otherwise redundant to the function of the model. 

Pruning for ANNs has a multitude of forms and methods~\cite{he2023structured}. The methods are either more static, independent of the model inputs, or more dynamic, dependent on the model inputs. The most static/independent methods are based on the weights of the actual model and can be performed without any external data, making them fast but not necessarily good for specific applications. Some partly dynamic/dependent methods run the data through the model only once but keep histories at specific points and then use that data to perform pruning. The most dynamic/fully-dependent methods take advantage of the training method of ANNs to have them prune their own weights. There are, of course, exceptions to each of these categories and exceptions to the categorization itself. 
Conventionally, pruning can be structured, unstructured, or a hybrid of the two. Structured means that the weights are not pruned individually, instead they are pruned in groups across the input vectors so that the output vector has fewer dimensions. Unstructured pruning instead does prune weights directly such that the output vector remains in the same format but becomes sparse. While both methods can speed up calculations, unstructured pruning that can handle sparse matrices often leads to a high compression rate
but requires specific hardware or library support for realistic compression~\cite{he2023structured}.

\begin{figure}
    \centering
    \includegraphics[width=\linewidth]{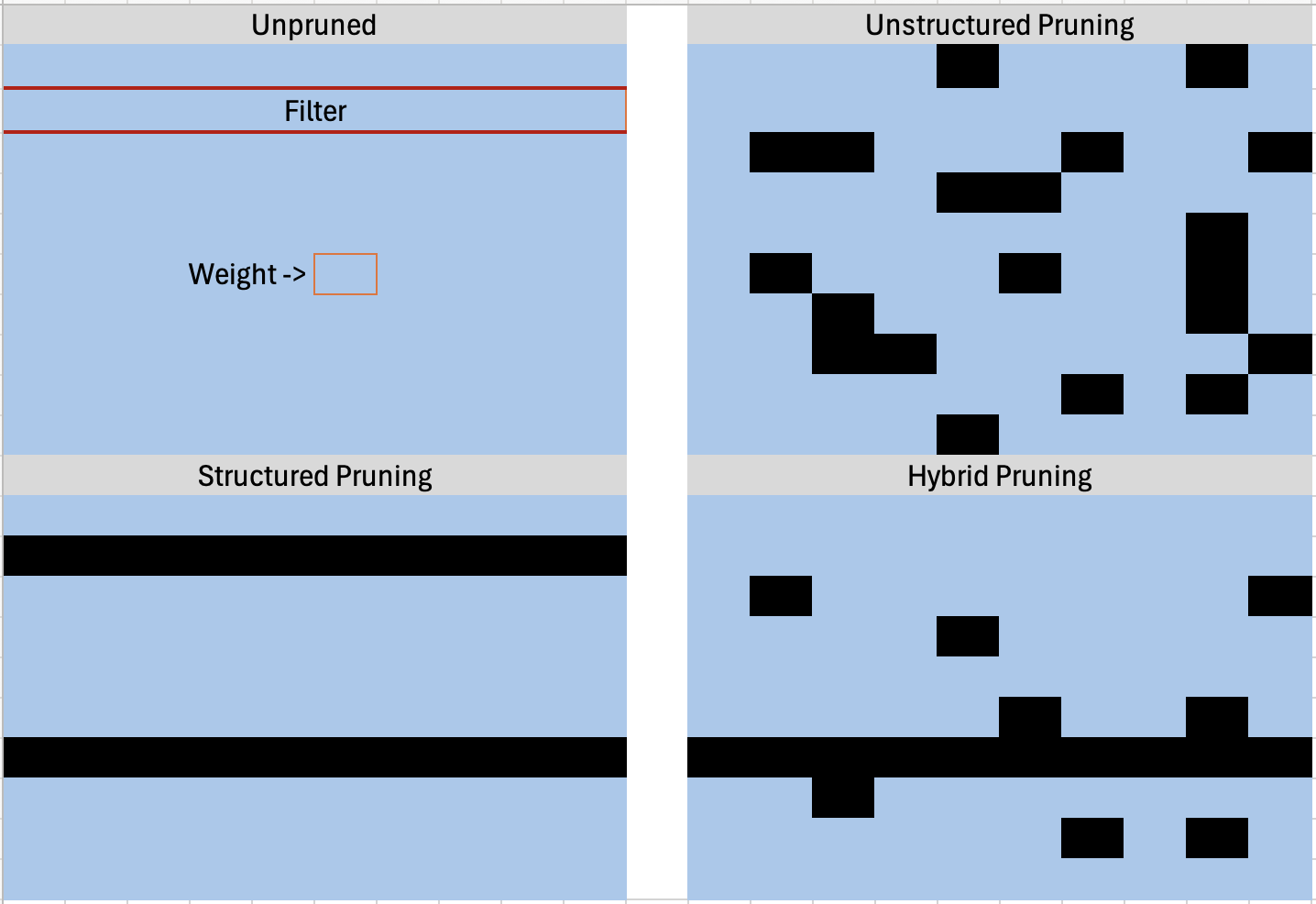}
    \caption{Filter shown with different pruning methods}
    \label{fig:struct_pruning}
\end{figure}

Figure~\ref{fig:struct_pruning} shows the difference between structured and unstructured pruning visually. The top left is an unpruned filter where each cell is a weight, columns is an input value from the features and each row will aggregate into a single value in the output vector. The bottom left is structured pruning where several output values are set to zero, visualized by darkening all of the weight cells associated with the pruned output values. The top right is unstructured pruning where individual weights are pruned but not whole columns or rows.

Finally, the bottom right shows hybrid pruning where both structured and unstructured pruning are being utilized.

One of the popular pruning approaches is to treat the absolute value of the weights of each filter as an importance measure to that filter.  It is a simple, standard\cite{L1document}, and static method for pruning. The higher weights are considered to be more important, or of more use to the model, and the lower weights to be less important, or less useful to the model~\cite{Distangling}. Therefore, less important weights are then pruned from the model.

Another static/independent method is random pruning, which works similarly to Dropout but is permanent instead of being regenerated on each new value and is used during inference time.

An example of a partly dynamic/dependent method would be if one was to use the gradient decent values from a single training epoch to calculate the importance of each weight. To then prune based off of that importance in the same manner as the weight based pruning. Fully dynamic methods are even more diverse than partly dynamic methods, they can utilize options such as modifying the loss function to account for a loss per non-zero node over a limit and count those nodes that become zero as pruned nodes. This method allows the model itself to determine importance in the same manner as picking weights.

\begin{table*}
    \centering
    \begin{tabular}{ccccc}
        \toprule
        Algorithm & Structure & Type & Method of identifying importance & \\
        \midrule
        ADMM\-joint & Hybrid & Dynamic/Fully-Dependent & Regularizer based & \\
        Bert\-Theseus & Structured & Dynamic/Fully-Dependent & Layer replacement based & \\
        DAIS & Structured* & Dynamic/Fully-Dependent & Regularizer based & \\
        Iterative\-Theseus & Structured & Static** & Layer replacement based & \\
        ThiNet & Structured & Partly Dynamic/Dependent & Heuristic measure & \\
        Random Pruning & Structured & Static & Random & \\
        Complete replacement & Structured & Static & Layer replacement based & \\
        \bottomrule
    \end{tabular}
    \caption{Algorithms used}
    \label{tab:algorithm-list}
\end{table*}

There are more pruning methods besides these examples though, we were only able to test out a few in our experiments that show a variety of categories. Specifically, we picked out a small assortment of high performing algorithms to test on collected from both~\cite{he2023structured} and through our own exploration of the literature.

The contribution of our comparison of these algorithms is that the model structure is outside of their original usage to see if these methods are widely generalize-able to network intrusion datasets and to compare them against the diverse competition.

\section{Algorithms}
\label{sec:Algorithms}

In our evaluation of pruning approaches for NIDS, we identified eight total algorithms or methods for implementation and testing using the ACI IoT dataset~\cite{ACIIOTdataset}, ADMM-joint~\cite{LEE2022107988}, Bert-Theseus~\cite{xu2020bert}, DAIS~\cite{guan2022dais}, Thinet~\cite{luo2017thinet}, Iterative-Theseus, Random pruning, and a complete retrain. Of these algorithms, ADMM and DAIS two are based on using regularizers during extra training of the model to perform pruning, the first of the two also uses hybrid pruning instead of structured pruning. The Bert-Theseus is a method of pruning whole layers instead of just individual filters. Thinet uses a heuristic to identify unneeded filters. 
We propose Iterative-Theseus as an update on Bert-Theseus.
The last two methods we have are just acting as a control, the random pruning should show how the model would behave if you randomly removed filters to reduce it under a quota and then retrained, and complete retrain just starts training fresh with a new model.

Each algorithm was chosen for a different purpose. ADMM-joint inspired the start of the project and as such must have been included in our tests. DAIS was chosen due to the results from the VGG-16 model on the Imagenet-1k dataset as seen in the survey paper. DAIS actually increased in accuracy by nearly $3$ percent after pruning, which seemed useful to test. Thinet was likewise chosen due to high accuracy in the survey, but with the added requirement that it had available compatible code with our testing method to be sure we were implementing it as well as possible. Thinet had an accuracy increase of $1.5$ after pruning. BERT-Theseus was a method that utilized methods in the direction we wanted to explore and had not been directly compared to the other algorithms before. Task Oriented Feature Distillation was chosen for a similar reason to DAIS and Thinet, but from \cite{gou2021knowledge} instead, reaching a $7$ percent increased accuracy on Resnet-50 after learning from a a teacher from Resnet-152 which is the best in the Offline Distillation category. Offline Distillation was chosen because it seemed least like BERT-Theseus which had already been selected, for the widest net to be cast of algorithms. As none of the other algorithms are related each other, it seemed best to select for a broad lineup for more uniform coverage rather than a deep dive into related methods. Then we wanted to get a good comparison of each so we decided to make a baseline, this was to just prune filters randomly from the model. When the random pruning did surprisingly well, we decided it might be a good idea to test against a model just trained with the data available in the pruning portion, which recreates a model to match the pruned number of parameters. 

\subsection{ADMM-joint} ADMM (Alternating direction method of multipliers)-joint~\cite{LEE2022107988} is an algorithm of retraining a model using binary masks for the output of each filter. This algorithm uses the ADMM method of allowing items to go beyond required bounds for bounded variables, but then using regularizers to bring them back into bounds. ADMM-Joint uses three regularizers for its goals. The first limits the number of weights using L0 normalization, which counts the non-zero values in the weight matrix. The second relates the actual applied mask to the idealized mask that only has values between 0 and 1, inclusively. And the third makes sure the idealized mask is not any value between 0 and 1, exclusively. Combined these three regulairizers create the conditions that the mask should be aiming for. The model is then trained while these regularizers are continuously updated and applied to form the final mask for the filters.

The 'joint' in the name is referring to the concept that this algorithm is performing both structured and unstructured pruning simultaneously. The weights are pruned by the L0 norm, which is an unstructured pruning, where each weight is considered individually. While the binary masks are structured pruning as they are cutting out whole filters at a time.

Note, for our implementation of ADMM we used much of the code from the provided ADMM github repository. However, we applied the masking after each linear layer while the original code applied it during the matrix multiplication of the weights. We performed a test and found that this does not affect the gradients of the result. 

\subsection{DAIS} DAIS (Differentiable Annealing Indicator Search)~\cite{guan2022dais} attempts to make a binary mask for channels that is able to be trained by the model itself. This is accomplished by allowing the channel pruning value to be continuous from 0 to 1. Then using a sigmoid function initially to allow back-propagation to find and train this mask layer. As the model progresses through training, the range is scaled by an annealing factor so that the sigmoid becomes closer to a binary classifier. Channels are specific to CNNs so we interpret them to mean to be working with filters in our fully connected model, the difference being the dimensionality of the output, channels having multiple dimensions per input and filters only having one.

DAIS also uses a combination of three regularizers to apply a loss value to the current distribution of the weights. Unfortunately one of them was for residual network blocks, which we did not have in our training model structure and the lasso method was not controllable which was not good for our tests so it was disabled much the same as the original paper. That only left the third regularizer for use in our study. Additionally DAIS was originally made for convolutional blocks, while we are using fully connected. 

To further complications with using DAIS in our tests, DAIS uses a particular training scheme introduced by DARTS (Differentiable architecture search)~\cite{liu2018darts} that involves alternately training just the masking layer on a speculative future of the weights and then returning to train the weights. The original DAIS may have used a differently modified version of this training but the paper says to check a reference sheet that was unable to be located for a specific breakdown of the training method. 

\subsection{Bert-Theseus} Bert-Theseus~\cite{xu2020bert} is a Knowledge Distillation adjacent method. Specifically, using the classifications identified by Gou \textit{et al.}~\cite{gou2021knowledge}\footnote{The survey does not actually have BERT Theseus classified}, it could be classified as a Feature-Based Self-distillation method. 

It works by creating a new replacement layer for a series of layers in the original model network, this replacement layer is then randomly fed training samples alongside the original pathway at a given percentage chance. This means that at the start, given the percentage $p$, $p\%$ samples go to the new replacement layer and $(1-p)\%$ go to the original layer. This allows the model weights time to adapt to the new model structure. The probabilities are either slowly increased or eventually just replaced with higher values over time, eventually reaching $100\%$ when the layer is replaced entirely.

For these tests, we wanted to reach a specific percentage of filters as such we defined the replacement layers to reach percentages. This is done by selecting modules in batches of $1/(average pruning percent)$ and defining them as a single replacement layer. This is not always very accurate to the pruning percentage, but does make some variation available in the resulting data.

When the model has already reached a stable point where it is not necessarily getting much better anytime soon, if you are to replace one of the layers, and only train that layer and a few layers around it, it may be possible to find a better local minimum then before if it exists.

\subsection{ThiNet} ThiNet~\cite{luo2017thinet} is a heuristic style pruning technique using the output of the layer after the current layer to find what filters to prune. This is accomplished by finding the expected output of the current layer over some training samples, then taking those values and going row by row feeding them into the next layer padded by zeros. These outputs are each summed up to produce a total activation for each filter in the current layer, as seen by the next layer. The filters are pruned in order of increasing activation as the lowest activation levels are not contributing much to the next layer's total activation.

ThiNet also generates a weight to be applied to each of the remaining weights to hasten the retraining of the reduced model. This weight is calculated by the activations of the layers that are being pruned from the model.

\subsection{Iterative-Theseus} 

This method is one of our contributions. It is related to the Bert-Theseus method, as it is a layer-replacement of the current model. However, instead of training the new layer alongside the old layer, only the new layer is used, and only one layer is replaced at a time. While this idea is simple and likely have been tried in other domains, we would like to test a brute force method such as this to evaluate how it holds up against the more nuanced methods. 

\subsection{Baselines}
We also have two baselines to compare against. The first is random structured pruning where random filters are disabled down to a percentage of the originally available filters on each layer. The second is Recreation run where the entirety of the weights and biases are reset and resized to be in line with what is available in the random structured pruning after the pruning occurs. The model is trained using just the pruning dataset, which is a smaller dataset then the original model but giving access to the whole dataset would give an advantage.

\section{Model and Data}
\label{sec:Setup}

\begin{figure}
    \centering
    \includegraphics[width=0.5\linewidth]{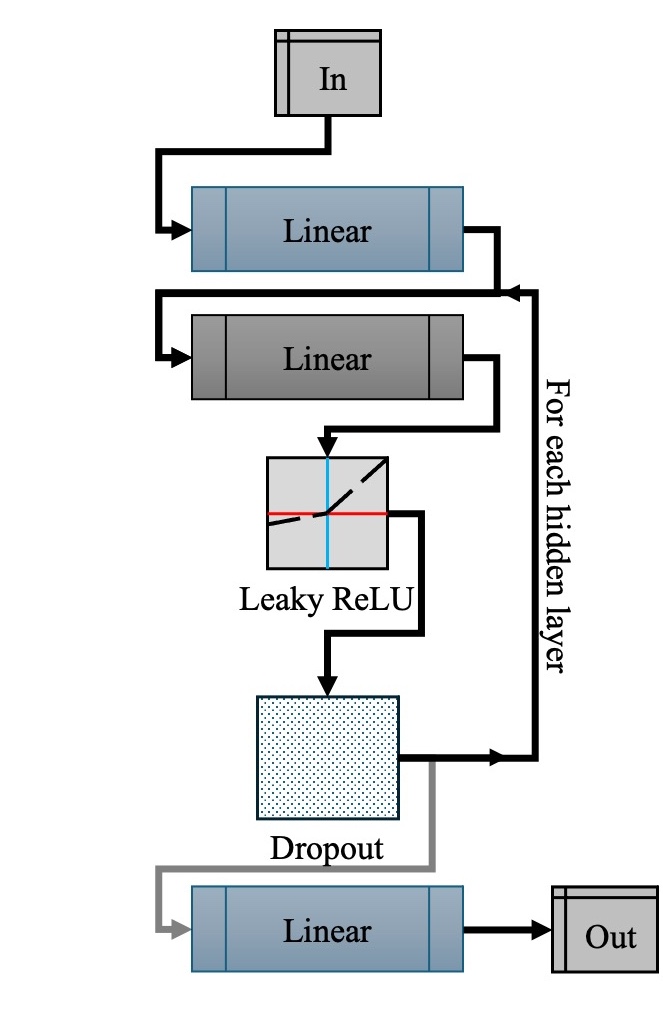}
    \caption{The model structure used for training 
    }
    \label{fig:modelstructure}

\end{figure}

\begin{table}
    \centering
    \begin{tabular}{ccccc}
        \toprule
          &  & Count &  & Grouped \\
          & Full & Under- & Grouped & Under- \\
        Classification & Count & sampled & Count & sampled\\
        \midrule
        Benign      & $601,868$      & $5,800$    & $601,868$  & $11,830$\\
        DNS Flood   & $18,577$       & $5,800$    & $20,816$*  & $11,830$*\\
        Dictionary Attack & $4,645$  & $4,645$    & $4,645$    & $4,645$\\
        Slowloris   & $2,974$        & $2,974$    & $2,974$    & $2,974$\\
        SYN Flood   & $2,113$        & $2,113$    & *         & *\\
        Port Scan   & $582$         & $582$     & $1,183$**  & $1,183$**\\
        Vulnerability Scan & $445$  & $445$     & **        & ** \\
        OS Scan     & $156$         & $156$     & **        & ** \\
        UDP Flood   & $68$          & $68$      & *         & *\\
        ICMP Flood  & $58$          & $58$      & *         & *\\
        \bottomrule
    \end{tabular}
    \caption{Counts of each class from the ACI dataset}
    \label{tab:ACI-counts}
\end{table}

In order to perform pruning on an artificial neural network, we need to define and train a network on a relevant dataset. We decided to use the dataset ACI-IOT-2023-payload~\cite{ACIIOTdataset} as our dataset due to its complexity, ease of access, and the direction we plan to go in in further work. We tried to use under-sampling, and grouping together some of the smaller classes to balance the dataset which is explored in Table~\ref{tab:ACI-counts}.

We used a simple model structure built from the \texttt{pytorch}~\footnote{https://pytorch.org} library consisting of a pair of linear layers surrounding $27$ fully connected linear layers with leaky ReLU activation and dropout. The number of layers was originally chosen by a short random search of layer sizes but was later reduced before the results as thinet's time scaled extremely poorly by the number of layers, which got in the way of ensuring the functionality of our testing codebase. We did not perform a longer search for a better model due to computational time required, but we left that for future work to compare against.

We used two different models, one with $175$ base filters per hidden layer that used the Under-sampled dataset\ref{tab:ACI-counts} and the other with $75$ filters per hidden layer using the Grouped Under-sampled dataset\ref{tab:ACI-counts} each with a batch size of $100$. The filters per hidden layer were augmented with an increase of $1$ per layer, ex. $175$, $176$, $177$, to aid in identifying problems in implementation. The grouped dataset combined several related classes into one to hopefully reduce loss for algorithms that are heavily dependent on loss values. These reduced numbers are marked in the table with a star or two stars for each group respectively.

The ACI-IOT-2023 dataset is a dataset that registers packet data from a group of devices as they receive a series of network attacks to identify network intrusions. Each entry in the dataset has srcip: source ip address, sport: source port number, dstip: destination ip address, dsport: destination port number, protocol\_m: the network protocol, sttl: source time to live, total\_len: length of network packet, payload: the contents of the network packet, stime: sending time, label: the target column. The dataset itself had a few modifications to prepare for the model. The stime column was removed because it might have had potential identifying information due to each class of network attack occurring at a different time, which times may not reflect in real world situations. The IP columns, ipv4, were broken into different integer columns from each section separated by bytes. Protocol was broken into single one-hot vectors, and the payload was broken into bytes and padded or truncated to 1500 columns/bytes. This ended up with $1515$ different feature columns and one target column. 

The models were initially trained with Crossentropy loss, the ADAM optimizer, a learning rate of $0.0009$ with a multiplicative scheduler of $0.1$ every $30$ epochs, $150$ total epochs, dropout at $0.0002$, the $27$ hidden layers each with $175$/$75$ filters. $80\%$ of the total dataset was used for training with the rest for testing the model's fit. For pruning, the number of epochs is decreased to $50$, for either the pruning loops if relevant or the retraining or both. The model was made up of linear layers with inter-spaced Leaky ReLU activation and dropout for training, seen in figure \ref{fig:modelstructure}. 

In the interest of being clear and open with the methods used in our work, our code and result data are available on Github\footnote{\url{https://github.com/ambroggi/Pruning-Project-for-Deep-Neural-Networks/tree/Before-Ontology-work}}.

\section{Results}
\label{sec:Results}
\begin{figure}
    \centering
    \includegraphics[width=1\linewidth]{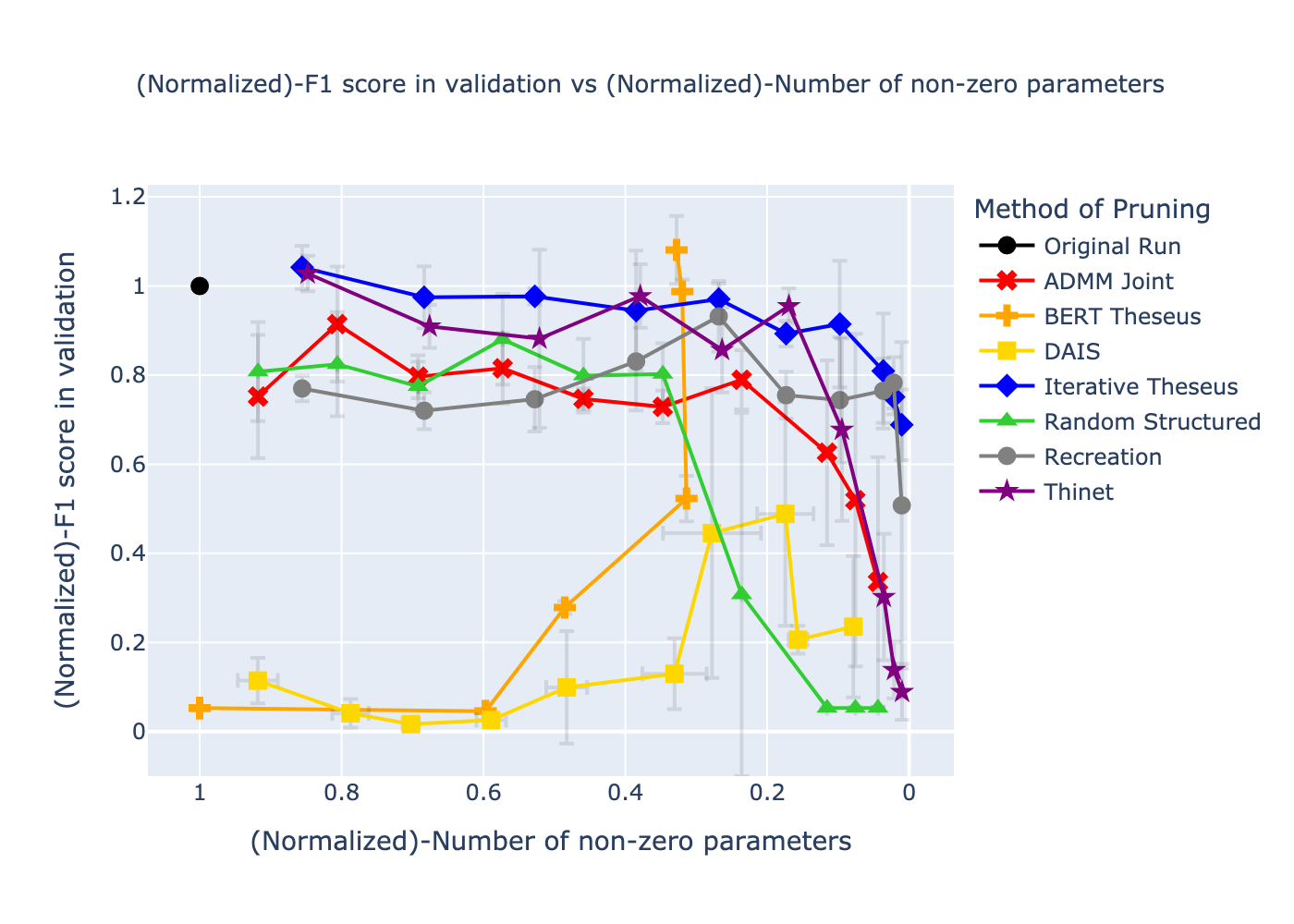}
    
    \caption{Big pruning model \textbf{scaled} such that the original run is at (1, 1)}
    \label{fig:big_model}
\end{figure}

We ran each algorithm three times on each of ten pruning percentages. We are defining 'pruning percentages' to be the estimated percentage of filters for each hidden layer after pruning has been applied, or an estimated replacement. For example, ADDM joint pruning has both 'k' and 'percent' options, so each is scaled by the pruning percent. Most of the algorithms are like this such that you can set a value per layer to a percentage or create a new layer with the the expected percent. The big exception to this is BERT Theseus which does not have any option to prune down by a specific percent, in this case we defined the blocks for the algorithm to be a size that is the multiplicative inverse to the average pruning percentage, due to the other constraint of it needing to be a whole number, the BERT Theseus tests do not have as much variation as the other tests in the lineup. We used two different groups of pruning percentages, from $0.99$ down to $0.12$ by increments of $0.12$-$0.13$ and from $0.12$ down to $0.04$ in increments of $0.04$ for finer increments near the end. This did result in us running the pruning percentage of $0.12$ twice, which we kept both versions of and combined them in the same method as the rest.

\begin{figure}
    \includegraphics[width=1\linewidth]{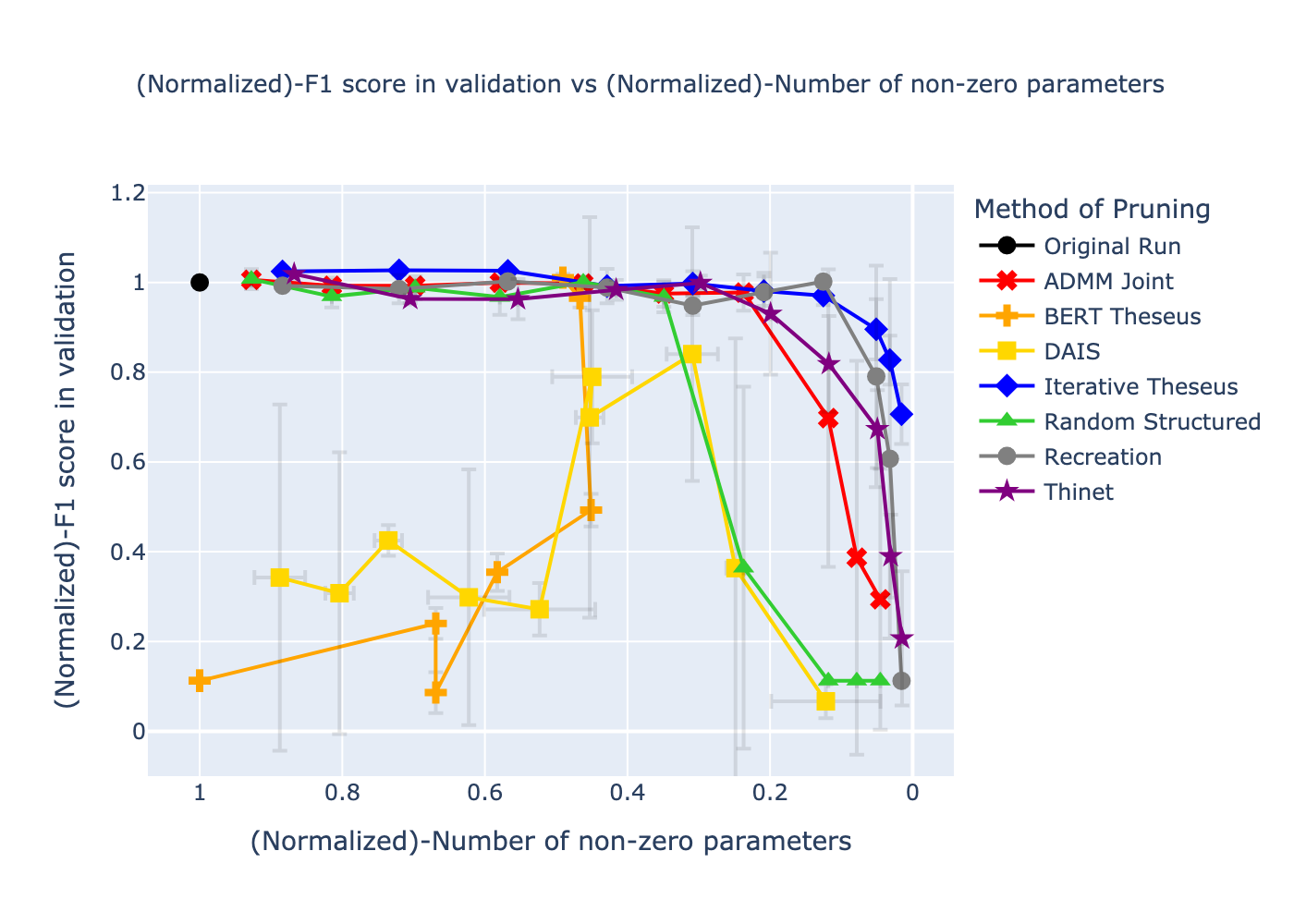}
    \caption{Smaller pruning model \textbf{scaled} such that the original run is at (1, 1)}
    \label{fig:small_model}
\end{figure}

To further be sure our findings were not by chance, we ran two separate tests with only minor changes to the model structure between them. Those changes being that the model seen in figure~\ref{fig:small_model}, has the hidden layers that are under half as large, the dataset grouped some related classes together resulting in fewer classes total, and the batch size is ten times as big as compared to the one seen in figure~\ref{fig:big_model}. The F1 scores themselves were calculated from the average F1 from each specific class in a single run.

You can see the results of our tests in Figures~\ref{fig:big_model} and \ref{fig:small_model}. They are made up of the original run, the black dot, the various pruning methods, seen as multi-color lines, and the retraining from scratch, the gray line. Each is made up of three runs that are individually scaled such that they are in respect to the original F1 score and number of filters. Then they are aggregated by means. This aggregation has error bars shown in very light gray calculated by $\frac{1.960*\sigma}{\sqrt{3}}$, given that there were three trial runs, we are aiming for a $95\%$ confidence interval hence a z score of $1.960$. In the equation $\sigma$ is the Standard Deviation, and we are assuming normal distribution. However, it may be incorrect to assume normal distribution for such a small sample size.

The first thing that appears out of the data is that two different styles appear, DAIS and BERT-Theseus appear to increase with more reduced models, we will refer to these as Helped algorithms, as in a more restricted prune helps them and the remaining pruning methods decrease after a certain point, we will call these Hindered algorithms. We will discuss any conclusions we can make about these categories later.

As the Hindered algorithms are more common and more expected, we will analyze their results first. Each hindered algorithm appears to have a point where they start decreasing in accuracy fast, this critical point appears to be related to the specific pruning algorithm being used, with Random Structured falling off first, followed by ADMM, Thinet, Recreation, and Iterative Theseus in that order. This is likely the point where the pruning method begins pruning critical nodes with no remaining redundancy which will be discussed later.

The best in terms of F1 before that point, as seen by both Figures, appears to be Iterative Theseus, however, these appearances do not tell the full story as Iterative Theseus takes by far the longest time to prune out of any of the Hindered algorithms and only takes more time as more is pruned from the model. The remaining algorithms take far less time to prune and are therefor likely more reasonable in application. The next best algorithm is Thinet, although it can be argued that the basic Recreation method is close. Thinet is on the slower side of the Hindered algorithms as a whole to prune, however it takes nearly an order of magnitude less time than Iterative Theseus. Additionally Thinet takes less time to prune the more pruning you perform due to an inverted pruning selection, picking out what to keep instead of what is gotten rid of.

Complete Recreation of the model is the next best, at least during the later stages of pruning, which hints that our initial model was likely made too large for the dataset. There appears to be a significant dropoff in the performance of Recreation in the bigger model seen in Figure~\ref{fig:big_model} at the very start, this may just be the results of a three times difference in the training time for the initial model but is something to keep an eye on. ADMM does not do as well as we were expecting it to do but is otherwise not very remarkable in our tests.

The worst of the Hindered category is Random Structured pruning. It starts off doing well, and even does better an half of the rest at one point, but is the first to hit the point of pruning critical nodes. Although this point appears to occur in a range as evidenced by the large error bar at just around $0.2\%$ where it both has and has not reached the critical point.

Of the two Helped algorithms, it appears that both of the Helped algorithms do not always perform adequately for the data
BERT-Theseus actually turns backwards at the end of each of its lines increasing in parameter count. This is just caused by our model structure and selection method of the pruned layers, given that the layer size increases and we are organizing layers into chunks, increasing the size of the chunk by a little bit does not increase the number of chunks but does increase the output dimension of the data, which results in an increased number of parameters. DAIS appears to be highly variable in its accuracy, with some results doing well and others doing poorly. It does tend to do better when the pruning rate increases but is very unstable while doing so.

\section{Conclusion}
\label{sec:Conclusion}
The first conclusion we feel that we can make is that our initial model was too large. We attempted to make a model that was able to be reasonably pruned and achieved a high F1 score. However the number of algorithms performing better at lower parameter counts and the long period before the dropoff of the Hindered algorithms indicate that the initial model may have individually done better if it was smaller in the beginning. Additionally, the good performance of the Recreation run also supports this conclusion. However, the model still can be pruned, although results may be skewed to be higher than expected.

The point where the Hindered algorithms turn downwards is likely the point where they begin pruning critical nodes with no backup redundancy, meaning that they begin to loose critical information. As a minimum number of nodes must be present for there to be enough width to distinctly identify classes, such as a single filter would not be able to distinctly identify the $10$ classes seen here, so there must be a point of too much pruning. This point does not seem to be set in stone, even within a single algorithm as evidenced by the comparatively large standard deviation for values around this critical point. We believe this large standard deviation appears due to some runs hitting the point earlier than the testing percentage and others later. 

The two categories we see in our results, those algorithms that are Hindered and Helped by more pruning, the Hindered algorithms we consider to be the default because it has the majority of the algorithms including the two baselines we are using. From this perspective, it is just the Helped algorithms that are performing unexpectedly. We believe that each of the two Helped algorithms are performing in this manner for different reasons. The BERT-Theseus method is performing poorly because at lower pruning rates, each module in the model is effectively being replaced by an equal module that is just untrained, that means the model has to train twice the number of modules since each layer relies on the layer before it. The extra training requirements do not help and just create confusion, we believe that extra training could bring the BERT-Theseus early models up to the point of Recreation but not much further than that. Once more layers can be removed though, there becomes far less to retrain and thus the model gets "Helped". DAIS however uses a modified the loss function which may have overtaken the normal loss applied to retrain the model, and focus too much on pruning. Perhaps applying gradient clipping would help DAIS work well on more lightly pruned models. However that is just speculation.

BERT-Theseus appears to do well when pruning by a lot and combining several layers together, however this may be due to the model having too many layers to start with. But even if that is the case it results in a better model overall which is good. Our conclusion is that BERT-Thesesus makes a good pruning method, but only if you are trying to remove large segments of the model.

DAIS is competitive with the other algorithms at lower weight pruning percents when it works, however it is unreliable for the entirety. Thus we do not feel like the algorithm is good when applied to Network Packet datasets, or non-residual models, which limit the amount of regularizers we can use. This might also be caused by the training cycle not being well described, perhaps a better training cycle would improve the reliability of the algorithm. Our conclusion with DAIS is that it needs to be further modified to work well on this problem.

ADMM works better than randomly pruning the model but appears to be more well designed for CNN models instead of purely fully connected models. This may have decreased its performance. Our conclusion with ADMM is that it works, but is not the best method of pruning in this domain.

The Iterative Theseus method we devised appears to do very well, but does not work fast. We believe that a brute force method like Iterative Theseus needs to be countered by a more optimal method of retraining the model over time, such that it can be run in reasonable time frames. 

The work of this paper has investigated a selection of five existing pruning and knowledge distillation methods and compared them on a novel dataset and model structure to analyze suitability for use with the dataset. From those tests we have found that the best algorithm for the ACI dataset on a fully connected neural network is Thinet due to higher performance and reasonable pruning time. Alternatively, completely replacing the layers to prune and retraining them also does well but takes much longer. We have also found that most analyzed pruning algorithms made for CNN networks cannot be transferred over to fully connected networks without losing significant potency. This is concerning due to the similarities between the structures, especially for future plans to apply the pruning methods to other model types.

\section*{Acknowledgment}

This work was supported by the U.S. Military Academy (USMA) under Cooperative Agreement No. W911NF-22-2-0160. The views and conclusions expressed in this paper are those of the authors and do not reflect the official policy or position of the U.S. Military Academy, or U.S. Army.

\balance

\bibliographystyle{IEEEtran}
\bibliography{references}

\appendix

\onecolumn
{
\small
\begin{center}
\begin{longtable}{lllrlrlrlr}
\toprule
 &  & \multicolumn{4}{l}{\underline{small}} & \multicolumn{4}{l}{\underline{big}} \\
 &  & Parameters & F1 Score & Standard & Time & Parameters & F1 Score & Standard & Time \\
PruningSelection & Percent &  & & Deviation & (Log) &  &  & Deviation & (Log) \\
\midrule
\endfirsthead
\toprule
 &  & \multicolumn{4}{l}{\underline{small}} & \multicolumn{4}{l}{\underline{big}} \\
 &  & Parameters & F1 Score & Standard & Time & Parameters & F1 Score & Standard & Time \\
PruningSelection & Percent &  & & Deviation & (Log) &  &  & Deviation & (Log) \\
\midrule
\endhead
\midrule
\multicolumn{10}{r}{Continued on next page} \\
\midrule
\endfoot
\bottomrule
\endlastfoot
Original Run & N/A & 329k & 0.945 & ±0.00 & 0.019* & 1233k & 0.769 & ±0.04 & 0.015* \\
\cline{1-10}
\multirow[c]{10}{*}{ADMM Joint} & 0.99 & 305k & 0.951 & ±0.01 & 0.023 & 1132k & 0.574 & ±0.09 & 0.020 \\
 & 0.87 & 268k & 0.938 & ±0.02 & 0.023 & 994k & 0.703 & ±0.11 & 0.020 \\
 & 0.74 & 230k & 0.939 & ±0.01 & 0.023 & 853k & 0.611 & ±0.01 & 0.020 \\
 & 0.62 & 190k & 0.945 & ±0.02 & 0.024 & 706k & 0.629 & ±0.10 & 0.020 \\
 & 0.49 & 152k & 0.943 & ±0.02 & 0.023 & 566k & 0.574 & ±0.04 & 0.020 \\
 & 0.37 & 115k & 0.922 & ±0.03 & 0.023 & 428k & 0.559 & ±0.00 & 0.019 \\
 & 0.25 & 78k & 0.924 & ±0.04 & 0.023 & 291k & 0.608 & ±0.09 & 0.019 \\
 & 0.12 & 38k & 0.660 & ±0.43 & 0.023 & 142k & 0.481 & ±0.22 & 0.019 \\
 & 0.08 & 25k & 0.366 & ±0.45 & 0.022 & 93k & 0.400 & ±0.31 & 0.018 \\
 & 0.04 & 14k & 0.278 & ±0.30 & 0.022 & 54k & 0.250 & ±0.22 & 0.018 \\
\cline{1-10}
\multirow[c]{10}{*}{BERT Theseus} & 0.99 & 329k & 0.106 & ±0.00 & 0.038 & 1233k & 0.041 & ±0.00 & 0.027 \\
 & 0.87 & 329k & 0.106 & ±0.00 & 0.038 & 1233k & 0.041 & ±0.00 & 0.027 \\
 & 0.74 & 329k & 0.106 & ±0.00 & 0.038 & 1233k & 0.041 & ±0.00 & 0.027 \\
 & 0.62 & 329k & 0.106 & ±0.00 & 0.038 & 1233k & 0.041 & ±0.00 & 0.027 \\
 & 0.49 & 220k & 0.227 & ±0.04 & 0.023 & 736k & 0.035 & ±0.00 & 0.016 \\
 & 0.37 & 220k & 0.081 & ±0.05 & 0.023 & 736k & 0.035 & ±0.01 & 0.016 \\
 & 0.25 & 192k & 0.335 & ±0.04 & 0.018 & 598k & 0.214 & ±0.00 & 0.013 \\
 & 0.12 & 148k & 0.466 & ±0.05 & 0.013 & 387k & 0.400 & ±0.04 & 0.009 \\
 & 0.08 & 153k & 0.912 & ±0.02 & 0.011 & 394k & 0.758 & ±0.02 & 0.009 \\
 & 0.04 & 161k & 0.956 & ±0.01 & 0.010 & 404k & \bfseries 0.828 & ±0.02 & 0.008 \\
\cline{1-10}
\multirow[c]{10}{*}{DAIS} & 0.99 & 292k & 0.324 & ±0.39 & 0.883 & 1132k & 0.087 & ±0.04 & 0.980 \\
 & 0.87 & 265k & 0.291 & ±0.32 & 0.967 & 971k & 0.031 & ±0.02 & 0.871 \\
 & 0.74 & 242k & 0.402 & ±0.04 & 0.923 & 866k & 0.013 & ±0.01 & 0.815 \\
 & 0.62 & 205k & 0.282 & ±0.29 & 0.905 & 726k & 0.020 & ±0.01 & 0.910 \\
 & 0.49 & 172k & 0.257 & ±0.06 & 0.904 & 595k & 0.073 & ±0.10 & 0.857 \\
 & 0.37 & 148k & 0.746 & ±0.15 & 0.906 & 407k & 0.102 & ±0.07 & 0.848 \\
 & 0.25 & 149k & 0.661 & ±0.46 & 0.883 & 342k & 0.345 & ±0.28 & 0.827 \\
 & 0.12 & 101k & 0.794 & ±0.37 & 0.878 & 215k & 0.371 & ±0.26 & 0.845 \\
 & 0.08 & 81k & 0.343 & ±0.52 & 0.885 & 193k & 0.159 & ±0.03 & 0.818 \\
 & 0.04 & 40k & 0.063 & ±0.04 & 0.894 & 96k & 0.184 & ±0.13 & 0.774 \\
\cline{1-10}
\multirow[c]{10}{*}{Iterative Theseus} & 0.99 & 291k & 0.968 & ±0.01 & 0.165 & 1055k & 0.800 & ±0.03 & 0.114 \\
 & 0.87 & 237k & \bfseries 0.971 & ±0.01 & 0.169 & 843k & 0.749 & ±0.07 & 0.228 \\
 & 0.74 & 187k & 0.970 & ±0.00 & 0.212 & 650k & 0.750 & ±0.03 & 0.285 \\
 & 0.62 & 141k & 0.938 & ±0.04 & 0.340 & 474k & 0.722 & ±0.07 & 0.348 \\
 & 0.49 & 101k & 0.943 & ±0.03 & 0.344 & 330k & 0.745 & ±0.03 & 0.463 \\
 & 0.37 & 68k & 0.927 & ±0.03 & 0.383 & 213k & 0.687 & ±0.06 & 0.530 \\
 & 0.25 & 41k & 0.918 & ±0.02 & 0.447 & 120k & 0.702 & ±0.11 & 0.589 \\
 & 0.12 & 16k & 0.847 & ±0.07 & 0.470 & 44k & 0.618 & ±0.11 & 0.712 \\
 & 0.08 & 10k & 0.782 & ±0.06 & 0.476 & 26k & 0.577 & ±0.03 & 0.786 \\
 & 0.04 & 5k & 0.668 & ±0.07 & 0.518 & 12k & 0.529 & ±0.07 & 0.787 \\
\cline{1-10}
\multirow[c]{10}{*}{Random Structured} & 0.99 & 305k & 0.952 & ±0.02 & 0.006 & 1132k & 0.618 & ±0.07 & 0.005 \\
 & 0.87 & 268k & 0.916 & ±0.02 & 0.006 & 994k & 0.636 & ±0.13 & 0.005 \\
 & 0.74 & 230k & 0.934 & ±0.03 & 0.007 & 853k & 0.594 & ±0.03 & 0.005 \\
 & 0.62 & 190k & 0.915 & ±0.04 & 0.007 & 706k & 0.677 & ±0.10 & 0.005 \\
 & 0.49 & 152k & 0.946 & ±0.01 & 0.007 & 566k & 0.613 & ±0.06 & 0.005 \\
 & 0.37 & 115k & 0.914 & ±0.03 & 0.007 & 428k & 0.618 & ±0.08 & 0.005 \\
 & 0.25 & 78k & 0.345 & ±0.41 & 0.007 & 291k & 0.235 & ±0.34 & 0.005 \\
 & 0.12 & 38k & 0.106 & ±0.00 & 0.007 & 142k & 0.041 & ±0.00 & 0.005 \\
 & 0.08 & 25k & 0.106 & ±0.00 & 0.007 & 93k & 0.041 & ±0.00 & 0.005 \\
 & 0.04 & 14k & 0.106 & ±0.00 & 0.007 & 54k & 0.041 & ±0.00 & 0.005 \\
\cline{1-10}
\multirow[c]{10}{*}{Recreation} & 0.99 & 291k & 0.938 & ±0.01 & 0.005 & 1055k & 0.592 & ±0.04 & 0.005 \\
 & 0.87 & 237k & 0.931 & ±0.03 & 0.006 & 843k & 0.552 & ±0.02 & 0.004 \\
 & 0.74 & 187k & 0.947 & ±0.00 & 0.005 & 650k & 0.572 & ±0.05 & 0.004 \\
 & 0.62 & 141k & 0.935 & ±0.01 & 0.006 & 474k & 0.635 & ±0.06 & 0.004 \\
 & 0.49 & 101k & 0.896 & ±0.02 & 0.006 & 330k & 0.715 & ±0.05 & 0.004 \\
 & 0.37 & 68k & 0.925 & ±0.04 & 0.006 & 213k & 0.579 & ±0.02 & 0.004 \\
 & 0.25 & 41k & 0.947 & ±0.01 & 0.006 & 120k & 0.568 & ±0.09 & 0.005 \\
 & 0.12 & 16k & 0.748 & ±0.32 & 0.006 & 44k & 0.586 & ±0.07 & 0.004 \\
 & 0.08 & 10k & 0.574 & ±0.41 & 0.006 & 26k & 0.603 & ±0.07 & 0.004 \\
 & 0.04 & 5k & 0.106 & ±0.00 & 0.005 & 12k & 0.380 & ±0.29 & 0.004 \\
\cline{1-10}
\multirow[c]{10}{*}{Thinet} & 0.99 & 286k & 0.963 & ±0.00 & 0.009 & 1045k & 0.789 & ±0.03 & 0.027 \\
 & 0.87 & 232k & 0.910 & ±0.01 & 0.008 & 833k & 0.700 & ±0.07 & 0.026 \\
 & 0.74 & 182k & 0.910 & ±0.04 & 0.008 & 642k & 0.673 & ±0.14 & 0.025 \\
 & 0.62 & 137k & 0.930 & ±0.02 & 0.008 & 467k & 0.750 & ±0.06 & 0.023 \\
 & 0.49 & 98k & 0.944 & ±0.02 & 0.007 & 325k & 0.660 & ±0.10 & 0.020 \\
 & 0.37 & 65k & 0.879 & ±0.14 & 0.007 & 208k & 0.733 & ±0.02 & 0.017 \\
 & 0.25 & 38k & 0.773 & ±0.11 & 0.007 & 116k & 0.523 & ±0.18 & 0.013 \\
 & 0.12 & 16k & 0.637 & ±0.11 & 0.006 & 43k & 0.233 & ±0.15 & 0.009 \\
 & 0.08 & 10k & 0.369 & ±0.10 & 0.006 & 26k & 0.105 & ±0.05 & 0.007 \\
 & 0.04 & 4k & 0.196 & ±0.15 & 0.006 & 12k & 0.067 & ±0.05 & 0.006 \\
\cline{1-10}
\caption{Number of Non-zero parameters, pruning time, and the F1 score in validation averaged over the three tests performed for each type of test, the standard deviation of these tests is also given. The pruning time is scaled so that the longest test took $1.00$ and then averaged over the three runs. Each type has an algorithm and a weight prune percent, here just called percent, however, the hidden layers have a slightly lower, ($10\%$ lower) pruning percentage due to how we set up our tests. \\ $*$Original run time includes initial training time
}
\label{tab:output_data}
\end{longtable}
\end{center}
}
\twocolumn

\end{document}